\documentclass[manuscript,screen]{acmart}

\usepackage{algorithmic}
\usepackage{graphicx}
\usepackage{wrapfig}
\usepackage{textcomp}
\usepackage{xcolor}
\usepackage{booktabs}
\usepackage{multirow}
\usepackage{csquotes}
\usepackage{placeins}
\usepackage{soul}
\usepackage{pifont}

\usepackage{xurl}

\newcommand{\cmark}{\ding{52}}
\newcommand{\xmark}{\ding{55}}
\newcommand{\ie}{{\em i.e., }}
\newcommand{\eg}{{\em e.g., }}

\newcommand\nvprof{\texttt{nvprof}}

\AtBeginDocument{%
  }

\setcopyright{cc}
\setcctype{by}
\acmJournal{TAISAP}
\acmYear{2026} \acmVolume{1} \acmNumber{1} \acmArticle{}
\acmMonth{1} \acmDOI{10.1145/3811923}
\acmBooktitle{ACM Transactions on AI Security and Privacy}

\begin{document}

\title{InferNet: Exploiting Aggregate GPU Profiles as Side-Channel for DNN Architecture Inference}

\author{Raja Hasnain Anwar}
\email{ranwar@umass.edu}
\affiliation{%
  \institution{University of Massachusetts Amherst}
  \country{USA}
}

\author{Jonah O'Brien Weiss}
\affiliation{%
  \institution{University of Massachusetts Amherst}
  \country{USA}
}

\author{Tiago Alves}
\affiliation{%
  \institution{Rio de Janeiro State University}
  \country{Brazil}
}

\author{Sandip Kundu}
\affiliation{%
  \institution{University of Massachusetts Amherst}
  \country{USA}
}

\author{Muhammad Taqi Raza}
\affiliation{%
  \institution{University of Massachusetts Amherst}
  \country{USA}
}

\begin{abstract}
Deep Neural Networks (DNNs) have become ubiquitous for their ability to solve problems across various domains, including computer vision, natural language processing, and speech recognition. However, as their adoption grows, they face a range of security threats, such as model stealing, architecture extraction, and manipulation, which can compromise their integrity, privacy, and functionality. Past works have relied on complex, fine-grained, and time-series analysis to launch DNN model extraction attacks. These approaches require extensive amounts of data, which are often challenging to acquire and analyze effectively. 
This paper introduces InferNet, an attack method that leverages simple, non-intrusive, and coarse-grained system-level information to identify the underlying DNN architecture of a victim's application. By analyzing GPU kernel calls, memory events, and system-level metrics, InferNet fingerprints the DNN and infers its architecture with very high accuracy. It can predict the architecture family (\eg Inception vs. BERT), as well as the architecture variant (\eg InceptionV1 vs. InceptionV3). 
The evaluation results demonstrate the effectiveness of InferNet across AI/ML frameworks (TensorFlow, PyTorch), different DNN types (vision, LLMs), and hardware platforms (\texttt{NVIDIA Tesla T4}, \texttt{NVIDIA Quadro RTX 8000}). The results show that InferNet achieves 100\% model extraction accuracy using only a partial GPU profile under various attack settings.
\end{abstract}



\maketitle

\section{Introduction}
The last decade has witnessed Deep Neural Networks (DNNs) become a cornerstone of AI and machine learning, especially when tasks demand high accuracy and mimic human-like cognitive processes. This proliferation is increasingly facilitated by Machine-Learning-as-a-Service (MLaaS) platforms, which have democratized access to AI capabilities.: image processing~\cite{convnets, deep_convnets}, speech recognition~\cite{speech_recognition}, natural language processing~\cite{brown2020language, devlin2018bert, vaswani2017attention}, and automotive driving~\cite{kato2015open, baidu, tesla, waymo}, and many more\cite{ramesh2021zero, singer2022make, silver2017mastering, silver2016mastering, harika2022review, xu2019explainable}. The economic scale of this transformation is immense; the global MLaaS market is projected to expand from approximately USD 44.2 billion in 2024 to over USD 1.2 trillion by 2034~\cite{mlaas}.
Trained DNN models are no longer mere software artifacts but are now considered invaluable intellectual property (IP) and core business assets. With training compute costs estimated to be in millions of dollars, the high barrier to entry and the immense value embodied in the trained model make them a highly attractive target for theft and replication~\cite{sun2023deepintellectualpropertyprotection}.

The architectural paradigm of AI development is undergoing a significant evolution, moving from bespoke, task-specific models toward large-scale, pre-trained \enquote{foundation models}~\cite{stanford2024aiindexcharts}.
These models are subsequently fine-tuned to perform a multitude of downstream tasks, a strategy that has become dominant across the industry.
This centralization of value creates a corresponding centralization of risk. A successful attack that compromises a single proprietary foundation model no longer affects just one application; it can undermine an entire ecosystem of services built upon it. Consequently, the security of foundation models has transcended being a mere operational concern to become a strategic imperative for the entire AI industry.

The concentration of high-value IP within MLaaS platforms has naturally led to the development of model extraction attacks, where an adversary queries a model to reverse-engineer its functionality without direct access to its weights or architecture~\cite{liang2025modelextractionattacksrevisited}. Historically, these attacks have relied on privileged access to the host system to monitor fine-grained, low-level resources such as memory, cache, or PCIe traffic~\cite{hackett2023pinchadversarialextractionattack}. However, such methods are often impractical in real-world public cloud environments for several reasons: first, they require access to system resources that are typically unavailable to a co-located user (such as in a public cloud environment); second, the extraction process is prone to overfitting on limited query data, diminishing the stolen model's ability to generalize; and third, the reliance on fine-grained observations makes these attacks highly susceptible to data distortion and noise, leading to inaccurate model reconstruction.

The paradigm shift toward fine-tuning pre-trained foundation models fundamentally alters the adversary's objective.
The focus has shifted to leveraging standard pre-trained models, trained on large, publicly available datasets, and then fine-tuning them for specific tasks (\eg by adding a few layers and training on smaller, task-specific datasets)~\cite{han2021pre, chen2020generative, radford2018improving}. This shift in development practices reduces the need to extract model parameters~\cite{dnn_fingerprinting, cryptanalytic_extraction, mobile_apps}, as the core features of these pre-trained models are already well-established. Motivated by this 2-phase AI training approach, we propose stealthy and non-intrusive attacks that can easily infer the underlying DNN \textit{architecture} (\eg GPT, BERT, or ResNet) used by the target applications. Our primary goals are: (1) to accurately identify the DNN architecture without requiring physical access or root privileges in deployments that expose profiling interfaces, (2) to ensure robust architecture prediction with sparse and noisy data, (3) to achieve high accuracy prediction even when the model is compressed into binary executable file, and (4) to evaluate transferability across hardware/framework settings.

Our empirical analysis demonstrates that aggregate GPU profiles effectively distinguish different DNN architectures, such as convolution-heavy vision models and transformer-based language models. This is primarily due to the specific kernel calls and system-level workloads that vary across DNN architectures, and their variants. From our study, we identify two key insights: first, each combination of DNN layer type and size corresponds to a specific GPU kernel call; second, the type of kernel calls, their execution times, and resource utilization form a unique fingerprint for DNN architectures.

This paper introduces \emph{InferNet}, an attack method that exploits statistical GPU performance metrics and workload behavior during the execution of a DNN model to identify its underlying architecture. InferNet identifies key GPU kernel calls and memory management events, along with their execution times, to create unique fingerprints for individual DNN layers. These metrics, combined with system-level data (such as memory usage, clock speed, and power consumption), are aggregated to profile the DNN model, and serve as a side-channel. By correlating the GPU profiles with known DNN architectures, InferNet accurately identifies both the DNN family (\eg Inception, Transformer) and its variant (\eg InceptionV1, InceptionV3).

InferNet adopts a temporally coarse but semantically rich profiling approach. Unlike prior work that depends on physical or low-level side-channels (\eg PCIe sniffing or EM traces), our attack directly leverages GPU kernel names and system metrics captured via {\nvprof} aggregate mode. While aggregate mode lacks kernel execution order (compared to time-series), it suppresses profile noises from GPU scheduling and internal resource contention, and therefore,  provides precise kernel identities and summary statistics that form a strong architectural fingerprint. This challenges the prevailing assumption that fine-grained traces are essential for DNN architecture inference, and highlights the attack surface exposed by aggregate GPU profiling available to the attacker.

\begin{table}[t]
\caption{Distribution of DNN architectures across PyTorch and TensorFlow for each architecture family used in the InferNet attack. VT = Vision Transformer.}
\centering
\begin{tabular}{@{}cc|cc@{}}
\toprule
\multirow{2}{*}{\textbf{Model Class}} & \multirow{2}{*}{\textbf{DNN Family}} & \multicolumn{2}{c}{\textbf{Number of Variants}} \\
& & {PyTorch} & {TensorFlow} \\
\midrule
\multirow{11}{*}{\textbf{Vision}} & AlexNet & 1 & -- \\ & ConvNeXt & 3 & 3 \\ & DenseNet & 4 & 3 \\ & EfficientNet & 11 & 8 \\ & Inception & 2 & 2 \\ & MnasNet & 4 & -- \\ & MobileNet & 3 & 3 \\ & ResNet & 9 & 3 \\ & SqueezeNet & 2 & -- \\ & ShuffleNet & 4 & -- \\ & VGG & 8 & 2 \\ \midrule \multirow{2}{*}{\textbf{VT}} & MaxVit & 1 & -- \\ & ViT & 4 & -- \\ \midrule \multirow{7}{*}{\textbf{LLMs}} & ALBERT & -- & 1 \\ & BART & -- & 1 \\ & BERT & -- & 1 \\ & Gemma & -- & 1 \\ & GPT & -- & 1 \\ & RoBERTa & 2 & -- \\ & XLM & 2 & -- \\ & XLM-RoBERTa & -- & 1 \\ \bottomrule \end{tabular} \label{table:combined_dnn} \end{table}

We evaluate the practicality and generalizability of InferNet across 90 diverse DNN architectures, including vision models, large language models (LLMs), and vision transformers. The architecture pool encompasses 21 DNN model families, each with multiple variants (see ~\autoref{table:combined_dnn}). These models are pre-trained and implemented in two major AI/ML frameworks: TensorFlow~\cite{abadi2015tensorflow} and PyTorch~\cite{paszke2019pytorch}. Experiments are conducted on two hardware platforms: the \texttt{NVIDIA Tesla T4} GPU on Google Colaboratory and the \texttt{NVIDIA Quadro RTX 8000} GPU on an Ubuntu server.

Our evaluation results show that InferNet is effective across multiple DNN types, AI/ML frameworks, and GPU hardware configurations. Key results includes DNN architecture extraction accuracy of: (i) 100\% using complete GPU profile, (ii) nearly 100\% using only the top 3 kernel features, (iii) 100\% on modified DNN architectures,  and (iv) 75\% across GPUs.
Compared to similar methods, InferNet achieves 100\% architecture extraction accuracy over a candidate set that is over four (4) times larger than the next best method (refer to \autoref{table:related_comparison} in the Related Work, \S\ref{sec:related_work}). 
\newline

\noindent
\textbf{Contributions.}\quad
To summarize, our work makes the following important contributions:

\begin{itemize}
    \item We identify aggregate GPU profiles---comprising kernel calls and system metrics---as a novel side-channel to fingerprint DNN architectures.
    \item We develop InferNet, a machine-learning-based attack method that uses GPU profiles to extract a wide range of DNN architectures, \eg vision, and large language models (LLMs).
    \item Our method can fully recover the model architecture even when the model has undergone extensive optimizations.
    \item Empirical evaluation shows that the complete GPU aggregate profile is not required for the prediction; instead, a few key features are sufficient for accurate DNN architecture extraction.
    \item We characterize the transferability of InferNet across GPU platforms and AI/ML frameworks, and show that heterogeneous training data can substantially improve cross-environment prediction.
\end{itemize}

\noindent
\textbf{Organization.}\quad The rest of the paper is organized as follows.  Section~\ref{sec:related_work} reviews related work on DNN architecture leakage. Section~\ref{sec:background} provides a technical overview of deep neural networks (DNNs) and their acceleration in GPU-enabled systems. In Section~\ref{sec:insights}, we present empirical observations on the mapping of DNN layers to low-level GPU kernel calls, which form the foundation for the design of InferNet. Section~\ref{sec:design} outlines the threat model and explains the methodology for extracting DNN architectures using GPU profiles. Sections~\ref{sec:eval} and~\ref{sec:analysis} evaluate the effectiveness of our attack across different challenging scenarios. In Section~\ref{sec:counter}, we discuss potential countermeasures to mitigate model extraction attacks. Finally, Section~\ref{sec:conclusion} summarizes our findings.
\section{Review of Related Works}
\label{sec:related_work}

The high-value, centralized landscape of modern AI development has given rise to a diverse spectrum of security threats against DNNs. These threats include adversarial attacks, which manipulate model inputs to cause misclassification; data poisoning and backdoor attacks, which corrupt the training process to embed malicious behavior; and privacy-violating attacks like membership inference, which aim to determine if a specific data point was used in the model's training set~\cite{he2020towards}.
Among these threats, a particularly insidious category is the Model Extraction Attack (MEA), where an adversary seeks to steal the model's architecture and weights~\cite{wang2024deep, liang2025modelextractionattacksrevisited}.
This section provides a comprehensive review of the literature on model extraction, with a particular focus on side-channel attacks that exploit unintentional information leakages from the underlying hardware and software systems to infer a model's confidential architecture.

\begin{table*}[t]
\caption{Comparison with related works, which also formulate architecture extraction as a multiclass classification problem.}
\centering
\begin{tabular}{@{}l|cccc@{}}
\toprule
\textbf{Related Work}                            & \textbf{Side-Channel}                    & \textbf{\begin{tabular}[c]{@{}c@{}}Candidate\\ Size\end{tabular}} & \textbf{\begin{tabular}[c]{@{}c@{}} Accuracy\end{tabular}} & \textbf{\begin{tabular}[c]{@{}c@{}}Aggregate\\ Data\end{tabular}} \\ \midrule
Yu et al.~\cite{deep_em}                    & Electromagnetic Trace                    & 15                                                                      & 100\%                                                                                 & {\color{green}\cmark}                                                           \\
Hong et  al.~\cite{cache_security_analysis} & Cache Timing Trace                       & 13                                                                      & 100\%                                                                                 & {\color{red}\xmark}                                                             \\
Patwari et al.~\cite{dnn_fingerprinting}    & Memory, CPU, and GPU Usage Trace         & 20                                                                      & 99\%                                                                                  & {\color{red}\xmark}                                                             \\
Kumar Jha et al.~\cite{deep_peep}           & Memory, Timing, Power, and GPU Kernels   & 15                                                                      & 100\%                                                                                 & {\color{red}\xmark}                                                             \\
Liu et al.~\cite{virtual_gpu}               & Adversarial GPU Kernel Execution Latency & 5                                                                       & 100\%                                                                                 & {\color{red}\xmark}                                                             \\
Xiang et al.~\cite{power_sc}                & Power Trace                              & 6                                                                       & 96.5\%                                                                                & {\color{green}\cmark}                                                           \\
\textbf{InferNet}                      & \textbf{GPU Kernel Profile}              & \textbf{90}                                                             & \textbf{100\%}                                                                        & \textbf{{\color{green}\cmark}}                                                  \\ \bottomrule
\end{tabular}
\label{table:related_comparison}
\end{table*}

\subsection{Escalated Attack Privileges}
The landscape of Deep Neural Network (DNN) model extraction is defined by a fundamental trade-off between an attack's capabilities and its required level of privilege. At one end of the spectrum lie attacks that achieve near-perfect model reconstruction but demand elevated, often impractical, access.
For instance, the Hermes~\cite{hermes} attack accomplishes lossless model recovery by snooping unencrypted PCIe traffic between the CPU and GPU. Similarly, DeepSniffer~\cite{deepsniffer} reconstructs architectural layouts by analyzing the volume of memory read/write operations from bus snooping. Other methods have demonstrated success by analyzing physical emanations; the CSI NN attack, for example, uses electromagnetic (EM) traces to reverse engineer the architecture and weights of networks on embedded devices~\cite{batina2019csi}. While formidable, the reliance of these techniques on direct hardware interaction or physical proximity renders them infeasible in typical cloud or Machine-Learning-as-a-Service (MLaaS) environments~\cite{wang2022demystifying}. Moreover, attacks like DeepPeep~\cite{deep_peep} and MoSConS~\cite{leaky_dnn} require access to the victim DNN during model training. The impracticality of these high-privilege methods in common cloud settings highlights the need for an attack vector that does not rely on physical access or strong adversary capabilities.

\subsection{Microarchitectural and Time-Series Side-Channels}
A more practical threat model involves exploiting microarchitectural side-channels in shared computing environments. These attacks typically rely on capturing a fine-grained, time-series trace of events to infer the sequence of a DNN's operations. Cache-timing attacks~\cite{cache_security_analysis, lyu2018survey, ge2018survey}, using underlying mechanisms like Prime+Probe or Flush+Reload, are a prominent example. Some, like Cache Telepathy~\cite{yan2020cache}, monitor cache hits on specific, shared DNN library functions to reconstruct the layer sequence.

A more novel approach, GANRED~\cite{ganred_cache}, captures the overall cache timing signature and uses a Generative Adversarial Network (GAN) to train a substitute model that mimics this signature, thereby avoiding a dependency on shared libraries. Other time-series attacks have targeted the GPU, such as Leaky DNN~\cite{leaky_dnn}, which exploits the timing penalties of GPU context-switching to infer layer composition and hyperparameters during the training phase. The critical vulnerability of these time-series methods is their fragility; they are susceptible to system noise and can be defeated by countermeasures that disrupt the temporal order of operations, such as randomized kernel scheduling and precision limiting~\cite{rendered_insecure, leaky_dnn}. The fragility of the time-series methods and unreliability of microarchitectural approaches create a clear need for a more robust side-channel that is resilient to system noise and temporal countermeasures.

Recent work further broadens the landscape of architecture leakage across side-channel modalities. Arefin and Serwadda~\cite{arefin2021deep} show that architecture-level information remains exposed even without direct physical access; Batina et al.~\cite{batina2021sca} synthesize reverse-engineering attacks on neural networks through side channels; and Malan et al.~\cite{malan2023enabling} demonstrate DVFS-based fingerprinting in edge inference services. These studies reinforce that architecture information can leak through diverse hardware/software observables. InferNet differs by focusing on aggregate GPU profiling and multiclass prediction over a substantially larger candidate pool.

\subsection{Binary-based Attacks}
In scenarios where the adversary has access to the model's binary or executable file (\eg compiled CUDA code, ONNX, or LiteRT), several binary-based model extraction techniques become feasible. Prior works have demonstrated architecture recovery through function call and cache tracing~\cite{liu2020ganred}, control-flow analysis~\cite{hua2018reverse}, or side-channel observations of compiled code~\cite{batina2019csi, krishna2019thieves}. Tools such as angr, Ghidra, and IDA Pro have also been used to disassemble binaries and recover model components when access to the model artifact is assumed. These approaches, however, typically require stronger adversarial capabilities, such as file system access, reverse engineering expertise, or dynamic instrumentation. In contrast, InferNet targets a weaker and more realistic threat model, where the attacker relies solely on aggregate GPU profiling data to infer the DNN architecture.

Our proposed attach method, InferNet, introduces a paradigm that circumvents the limitations of existing approaches by strategically sacrificing temporal data for statistical robustness.
Attacks like Hermes~\cite{hermes}, DeepSniffer~\cite{deepsniffer}, and DeepPeep~\cite{deep_peep} assume identical conditions in both training and deployment phases of the DNN lifecycle. Although identical settings maximize performance for all attacks, InferNet retains significant efficacy under varied conditions, demonstrating its practical versatility. It exhibits broader applicability, encompassing a wider range of deep neural network architectures, including vision models, vision transformers, and large language models (LLMs). As shown in \autoref{table:related_comparison}, InferNet achieves 100\% architecture extraction accuracy across a diverse candidate set of 90 models, encompassing vision models, vision transformers, and large language models (LLMs) across both PyTorch and TensorFlow---a pool more than four times larger than the next-best classification-based method. 

\section{Background}
\label{sec:background}
This section provides the technical background on deep neural networks (DNNs) and their use of multithreading technology on modern GPUs. We begin by describing the structure of DNNs and their key building blocks. Next, we introduce the CUDA toolkit~\cite{cuda} and explain how it accelerates data processing on GPUs. Finally, we discuss how profilers can help monitor various GPU workloads involved in a DNN execution.

\subsection{Deep Neural Networks} 

Deep neural networks (DNNs) comprise a hierarchical organization of connected layers to model the relationship between the input $x$ and output $y$ by approximating an unknown function $f^\theta(x) = y$. Here, $\theta$ denotes the \emph{learned} parameters, \ie weights, biases, and other model-specific parameters of the layers in the architecture $f$; and $f^\theta$ is the resultant model. Computationally, the input $x$ passes through the DNN in a series of feed-forward matrix operations. Each layer processes the input to extract specific features, progressively transforming it into higher-level representations. The final output $y$ is typically a probability distribution over multiple categories.

DNNs operate in two main phases: training and inference. In the \textit{training} phase, the model processes input data to produce an output, which is compared to the ground truth using a loss function to quantify the error in the model's predictions. The model's parameters $\theta$ are adjusted iteratively through backpropagation to minimize this error~\cite{lecun2015deep}. \emph{Hyper-parameters}, such as learning rate, epochs, and batch size that control the training process, are independent of the data~\cite{lecun2015deep}. In the \textit{inference} phase, the trained model makes predictions on new, unseen data. When a DNN is deployed in a real-world setting, it operates in the inference phase.

\subsubsection{DNN Architecture}
DNN architecture is defined by its layers, their types or dimensions, and the topology of connections, which form a computational graph. Layers can be connected either sequentially or non-sequentially. In a sequential connection, each layer directly takes the output of the previous layer as input. In contrast, non-sequential connections include more complex structures such as skip connections, branches, or shared layers.

Modern DNN models can be broadly categorized into two types based on their application domains: \textit{vision models}, designed for processing image data, and \textit{large language models} (LLMs), specialized for understanding and generating text. Vision models typically employ architectures that process spatial data, utilizing layers such as convolutional and pooling layers to capture local patterns and hierarchies in images. These models rely on grid-like structures to preserve spatial relationships between pixels. In contrast, LLMs are designed to process sequential data, leveraging architectures like transformers~\cite{vaswani2017attention} with self-attention to capture dependencies across text tokens. While vision models focus on spatial feature extraction, LLMs emphasize understanding contextual relationships in sequential data. Therefore, the architecture and layer configuration of a DNN must be tailored to the specific characteristics of the data it is intended to process.

\subsubsection{DNN Optimization}
Modern `deep' neural networks (DNN) tend to have a large number of layers. It is, therefore, essential to reduce the resource requirements for such architectures without affecting their performance. DNN optimization techniques, like \textbf{pruning}~\cite{seminal_prune}, reduce the computation overhead by removing the least important parameters $\in \theta$. This creates a sparser architecture and has been shown to significantly decrease the size of the model with minimal impact on accuracy~\cite{seminal_prune}. Similarly, \textbf{neural architecture search (NAS)}~\cite{elsken2019neural} automates discovering optimized architectures tailored to specific constraints, \eg hardware and cost.
Depending on specific constraints, DNN architectures can be adapted into optimized variants, such as MobileNetV2~\cite{sandler2018mobilenetv2} and MobileNetV3~\cite{howard2019searching}, designed explicitly for resource-constrained hardware platforms.

\subsection{GPU-Accelerated DNNs}
Modern AI/ML frameworks leverage the parallel processing capabilities of GPUs~\cite{oh2004gpu} for DNN workloads, such as matrix multiplications and convolutions. GPUs come with advanced development toolkits, such as NVIDIA's CUDA~\cite{cuda}, which offer high-level APIs and libraries for general-purpose use.

\begin{wrapfigure}{r}{0.5\linewidth}
    \centering
    \vspace{-0.4cm}
    \includegraphics[width=\linewidth]{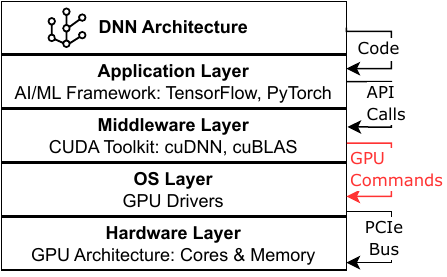}
    \caption{The architecture stack that runs a DNN on an NVIDIA GPU using CUDA. Red color indicates data collected for profiling.}
    \label{fig:systemstack}
    \vspace{-0.4cm}
\end{wrapfigure}

\subsubsection{CUDA}
CUDA is a parallel computing platform equipped with a suite of tools to create and optimize programs to run on NVIDIA GPUs. It acts as an interface between AI/ML frameworks, such as TensorFlow~\cite{abadi2015tensorflow} and PyTorch~\cite{paszke2019pytorch}, and GPU hardware. The  AI/ML frameworks define the DNN architecture and computational workflows, while CUDA translates high-level mathematical operations (\eg convolutions) into low-level GPU functions, called \textbf{kernels}\footnote{A CUDA kernel, not to be confused with an OS kernel, refers to a function executed concurrently by multiple GPU threads.}.
\autoref{fig:systemstack} depicts the key elements involved in the GPU system stack to run a DNN using the CUDA toolkit.

CUDA maps DNN operations, represented as a computation graph, onto the GPU hardware using specialized libraries. In CUDA libraries, each \textit{kernel} is designed to break down complex operations into smaller, unit-level tasks that can be executed concurrently across multiple GPU cores. For instance, matrix multiplication operations in fully connected layers use the cuBLAS (CUDA Basic Linear Algebra Subprograms)~\cite{cublas} library, while convolutional operations are optimized using the cuDNN (CUDA Deep Neural Network)~\cite{cudnn} library. Activation functions like ReLU and sigmoid are executed as element-wise operations using highly parallelized CUDA kernel calls.

\subsubsection{NVIDIA Profiler}

{\nvprof}~\cite{nvprof} is a command-line tool for monitoring CUDA activities during the execution of GPU-accelerated applications.
{\nvprof} captures details like the type of kernel executed and time spent on execution.
It also tracks memory-related activities, such as host-to-device and device-to-host data transfers, as well as memory allocation and deallocation events. Additionally. It monitors system-level performance metrics, including GPU clock speed, power consumption, and thermal data, to provide a comprehensive view of the application's performance. All the collected data is compiled into a detailed report, referred to as a \textit{profile}.

{\nvprof} can be configured to generate profiles in either time series or aggregate mode. In time-series mode, it records detailed temporal data, including the execution sequence of GPU kernels, memory operations, and performance metrics associated with each kernel invocation. It is, therefore, useful for analyzing fine-grained sequences of operations.

\begin{wrapfigure}{r}{0.45\linewidth}
\centering
\vspace{-0.45cm}
\includegraphics[width=\linewidth]{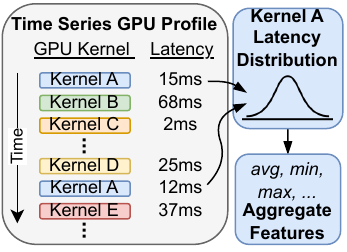}
\vspace{-0.75cm}
\caption{Collection of aggregate GPU profiles without preserving the kernel execution order.}
\label{fig:agg_features}
\vspace{-0.5cm}
\end{wrapfigure}

Unlike time-series, the aggregate provides a high-level view of resource usage and performance distribution. Here, {\nvprof} collects metrics including total time spent, number of calls, and average, minimum, and maximum execution times for each type of kernel. \autoref{fig:agg_features} illustrates the process of collecting and summarizing various metrics to create an aggregate GPU profile. \autoref{tab:profile} shows partial aggregated GPU profiles generated by {\nvprof} for 10 inferences on GoogleNet~\cite{szegedy2015going} and RoBERTa~\cite{liu2019roberta}. Similarly, it collects average, minimum, and maximum values for various system-level metrics (see \autoref{tab:system_signals}).  In this paper, we will only focus on the profiles collected in the aggregate mode of {\nvprof}.

\begin{table*}[t]
\small
\caption{Partial view of aggregated profiles acquired by running 10 inferences of GoogleNet~\cite{szegedy2015going} and RoBERTa~\cite{liu2019roberta} models on an NVIDIA Tesla T4 GPU using PyTorch framework. Full profile data would include $\sim$15 GPU kernels, $\sim$30 API calls, and 5 system signals.
}
\centering
\begin{tabular}{@{}cllcccccc@{}}
\toprule
\textbf{Model} & \textbf{Type} & \textbf{Name} & \textbf{Time(\%)} & \textbf{Time(ms)} & \textbf{\# Calls} & \textbf{Avg($\mu$s)} & \textbf{Min($\mu$s)} & \textbf{Max(ms)} \\ \toprule
\multirow{6}{*}{\rotatebox[origin=c]{90}{GoogLeNet}} & \multirow{3}{*}{Kernel} & cudnn\_infer\_volta\_scudnn\_winog...  & 31.87 & 13.94 & 190 & 73.36 & 20.74 & 0.26 \\
                            & & volta\_sgemm\_32x32\_sliced1x4\_nn    & 14.43 & 6.31 & 190 & 33.22 & 19.36 & 0.05\\
                            & & {[}CUDA memcpy HtoD{]}                & 8.87 & 3.88 & 354 & 10.96 & 0.64 & 0.75\\ \cmidrule(lr){2-9}
& \multirow{3}{*}{API call} & cudaDeviceGetStreamPriorityRange        & 39.73 & 232.00 & 1141 & 203.33& 0.39 & 231.07\\
                        & & cudaLaunchKernel                          & 33.93 & 198.11 & 2370 & 83.59 & 4.71 & 50.05 \\
                        & & cudaGetDevice                             & 3.78 & 22.09 & 28307 & 0.78 & 0.29 & 6.15\\ \midrule
                        
\multirow{6}{*}{\rotatebox[origin=c]{90}{RoBERTa}} & \multirow{3}{*}{Kernel} & [CUDA memcpy HtoD]  & 49.92 & 285.88 & 302 & 946.61 & 0.54 & 42.87 \\
                            & & volta\_sgemm\_128x64\_tn    & 29.67 & 169.89 & 720 & 235.96 & 81.28 & 0.44 \\
                            & & {volta\_sgemm\_128x128\_tn}  & 15.11 & 86.52 & 240 & 360.52 & 262.05 & 0.49 \\ \cmidrule(lr){2-9}
& \multirow{3}{*}{API call} & cudaLaunchKernel        & 53.20 & 942.32 & 3250 & 289.94 & 6.93 & 783.65\\
                        & & cudaMemcpyAsync         & 17.10 & 302.90 & 302 & 1002.99 & 5.12 & 43.07 \\
                        & & cudaDeviceGetStreamPriorityRange & 11.65 & 206.25 & 1 & 206249.62 & 206249.62 & 206.25 \\ \bottomrule
\end{tabular}
\label{tab:profile}
\end{table*}

\section{Aggregate GPU Profiling} 
\label{sec:insights}
We demonstrate that adversaries can infer DNN architectures by leveraging \textit{aggregate GPU profiles} as a side-channel. While these profiles are simple, non-intrusive, and coarse-grained (in contrast to time-series), they are powerful enough to reveal DNN architectures. Our work stands apart from previous works~\cite{deepsniffer, hermes, deep_peep} that rely on a complex analysis of fine-grained time-series data to reconstruct DNN architectures layer-by-layer with a certain degree of uncertainty. This section highlights key insights into the \textit{aggregate} CUDA kernel calls that reveal unique characteristics of DNN architectures.

\subsection{Mapping CUDA Kernels to DNN Layers}
By mapping CUDA kernel functions to DNN layers, it is possible to identify key characteristics of DNN architectures.
For example, matrix multiplication in fully connected layers is executed by dividing the input matrix into smaller blocks and assigning them to GPU threads for parallel computation. Similarly, convolutional layers use kernels that distribute the workload across threads to compute multiple filters simultaneously. NVIDIA's cuDNN library provides tailored functions to optimize operations like convolutions, where the computation strategy may vary based on filter size, \eg $3 \times 3$ vs. $5 \times 5$ convolutions. As a result, convolution operations with different filter sizes invoke different kernels.

\begin{wrapfigure}{r}{0.55\linewidth}
\centering
\vspace{-0.5cm}
\includegraphics[width=\linewidth]{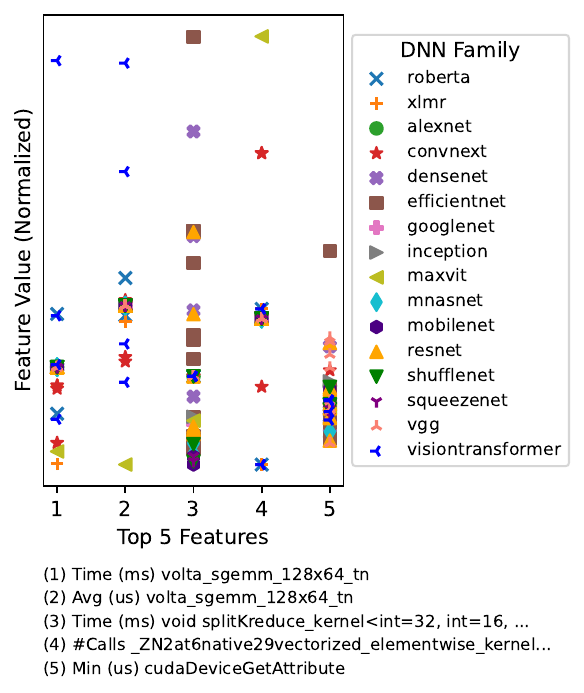}
\vspace{-0.75cm}
\caption{GPU kernel metrics for all 56 TorchVision and 4 TorchText models grouped by their architecture families. The features are ranked from the set of non-memory related GPU kernels using Recursive Feature Elimination on a Random Forest~\cite{breiman2001random} model.}
\label{fig:top_features}
\vspace{-0.8cm}
\end{wrapfigure}

Specifically, kernels like \texttt{volta\_sgemm\_AxB\_tn}, which handle matrix multiplications, can indicate the presence of fully connected layers and their relative sizes. Similarly, kernels such as \texttt{void splitKreduce\_kernel<int=A, int=B, ...>}\footnote{Truncated long kernel name.} provide insights into convolutional operations, including the size and the number of filters. Analyzing low-level kernel calls across multiple iterations can reveal and fingerprint architectural nuances specific to different DNN families. \autoref{fig:top_features} highlights the most common kernels for various DNN families. These kernels, as we ranked them by relevance, reveal the architectural characteristics that distinguish vision models from LLMs, as well as their types.

\subsection{Profiling CUDA Activities with \nvprof}
Aggregate profiling offers high-level summary metrics for kernel activities and resource utilization that can fingerprint DNN architectures.
\autoref{tab:system_signals} compares aggregated kernel and API metrics for two popular DNN models: GoogLeNet~\cite{szegedy2015going} and RoBERTa~\cite{liu2019roberta}. A closer examination of the metrics shows that GoogLeNet exhibits higher usage of convolutional kernels like \texttt{cudnn\_infer\_volta\_scudnn\_winograd\_128...}\footnote{Truncated long kernel name.}, accumulating over 46.3\% of all operations. The higher usage of convolution kernels aligns with the architectural characteristics of the vision models. In contrast, RoBERTa relies more heavily on specialized fully connected operations with kernels like \texttt{volta\_sgemm\_XXXXX\_tn}, reflecting the matrix-heavy computations needed for large transformer blocks in language models~\cite{vaswani2017attention}.

Additionally, system-level signals from \autoref{tab:system_signals} further emphasize the architectural differences between LLMs and vision models. GoogLeNet's lower average SM clock speeds and power consumption reflect its smaller size and fewer trainable parameters, \ie \(\sim \)6M~\cite{szegedy2015going}. Whereas, RoBERTa exhibits significantly higher power usage and clock rates, corresponding to its larger model size and a much higher number of trainable parameters, \ie \(\sim \)355M~\cite{huggingface}.

\begin{table}[t]
\caption{System-level signals acquired by running 10 inferences of GoogleNet~\cite{szegedy2015going} and RoBERTa~\cite{liu2019roberta} models on an NVIDIA Tesla T4 GPU using PyTorch framework.
}
\centering
\begin{tabular}{@{}clccccc@{}}
\toprule
\textbf{Model} & \textbf{Signal} & \textbf{Count} & \textbf{Avg} & \textbf{Min} & \textbf{Max} \\ \toprule
\multirow{4}{*}{\rotatebox[origin=c]{90}{GoogLeNet}} & SM Clock (MHz) & 13 & 563.1 & 300 & 585 \\
 & Mem Clock (GHz) & 13 & 4.6 & 0.4 & 5.0 \\
 & Temperature (C) & 26 & 51.5 & 51 & 52 \\
& Power (W) & 26 & 27.9 & 10.7 & 31.8 \\ \midrule
\multirow{4}{*}{\rotatebox[origin=c]{90}{RoBERTa}} & SM Clock (MHz) & 31 & 751.9 & 300 & 1125 \\
 & Mem Clock (GHz) & 31 & 4.8 & 0.4 & 5.0 \\
 & Temperature (C) & 61 & 62.1 & 61 & 63 \\
 & Power (W) & 60 & 34.5 & 11.8 & 56.6 \\ \bottomrule
\end{tabular}
\label{tab:system_signals}
\end{table}

\begin{figure}[t]
\centering
\includegraphics[width=0.65\linewidth]{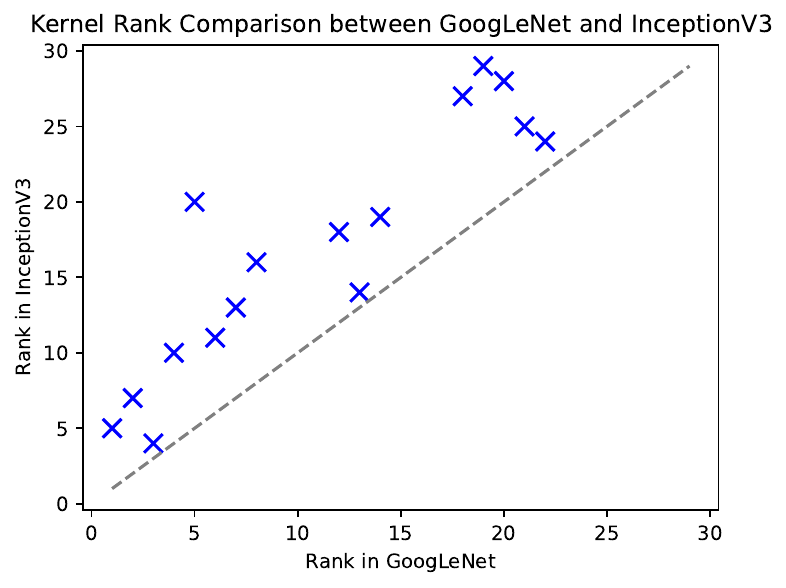}
\caption{Rank comparison of kernels used in both GoogLeNet and InceptionV3 models in PyTorch. Points further from the diagonal indicate greater differences in kernel usage between the models.}
\label{fig:inception_comparison}
\end{figure}

Within the same DNN family, variations in activity patterns emerge due to differences in layer depth and applied optimization strategies.  For example, GoogLeNet~\cite{szegedy2015going} (InceptionV1) and InceptionV3~\cite{szegedy2016rethinking} are both vision models that employ branched convolutions but differ in kernel configurations and architectural complexity. In GoogLeNet, a large portion of GPU time is spent on Winograd convolution~\cite{lavin2016fast} kernels, with \texttt{cudnn\_infer\_volta\_scudnn\_winograd\_128x...}\footnote{Truncated long kernel name.} accounting for 31.87\% of total execution time.  Other prominent operations include matrix multiplications using \texttt{volta\_sgemm\_32x32\_sliced1x4\_nn} (14.43\%) and host-to-device memory transfers (8.87\%). In contrast, the InceptionV3 profile reveals a significant portion of execution time spent in FFT-based operations like \texttt{fft2d\_r2c\_32x32}, which contributes 39.36\% of execution time, and \texttt{gemv2T\_kernel\_val}, a matrix-vector multiplication kernel, accounting for 22.24\%. 
\autoref{fig:inception_comparison} shows a scatter plot comparing kernel runtime ranks for both models. The ranking differences indicate that InceptionV3 employs more specialized and optimized kernels in place of the more generic kernels used in GoogLeNet. These distinctions in kernel usage and time distribution yield unique execution profiles, enabling fingerprinting of model variants even within the same architecture family.

\subsection{Key Takeaways}

Our analysis demonstrates that aggregate profiling effectively captures distinct execution characteristics of DNN families, such as convolution-heavy vision models or transformer-based language models. For instance, the divergence in kernel usage and system-level metrics between GoogLeNet and RoBERTa underscores the utility of aggregate GPU profiles in identifying architectures across diverse families. Similarly, the higher runtime contribution of FFT-based kernels in InceptionV3 compared to GoogLeNet highlights differences across variants of the same DNN family captured by GPU profiles. 
We argue that aggregate GPU profiles can be used as a side-channel for extracting DNN architectures. Our argument is based on two key observations:

\begin{enumerate}
    \item Each unique combination of DNN layer type and size maps to a specific CUDA kernel call.
    \item Patterns of kernel calls, along with their execution times and resource utilization, provide unique fingerprinting for DNN architectures.
\end{enumerate}
\section{InferNet Design}
\label{sec:design}
\begin{figure*}[t]
\centering
\includegraphics[width=0.9\textwidth]{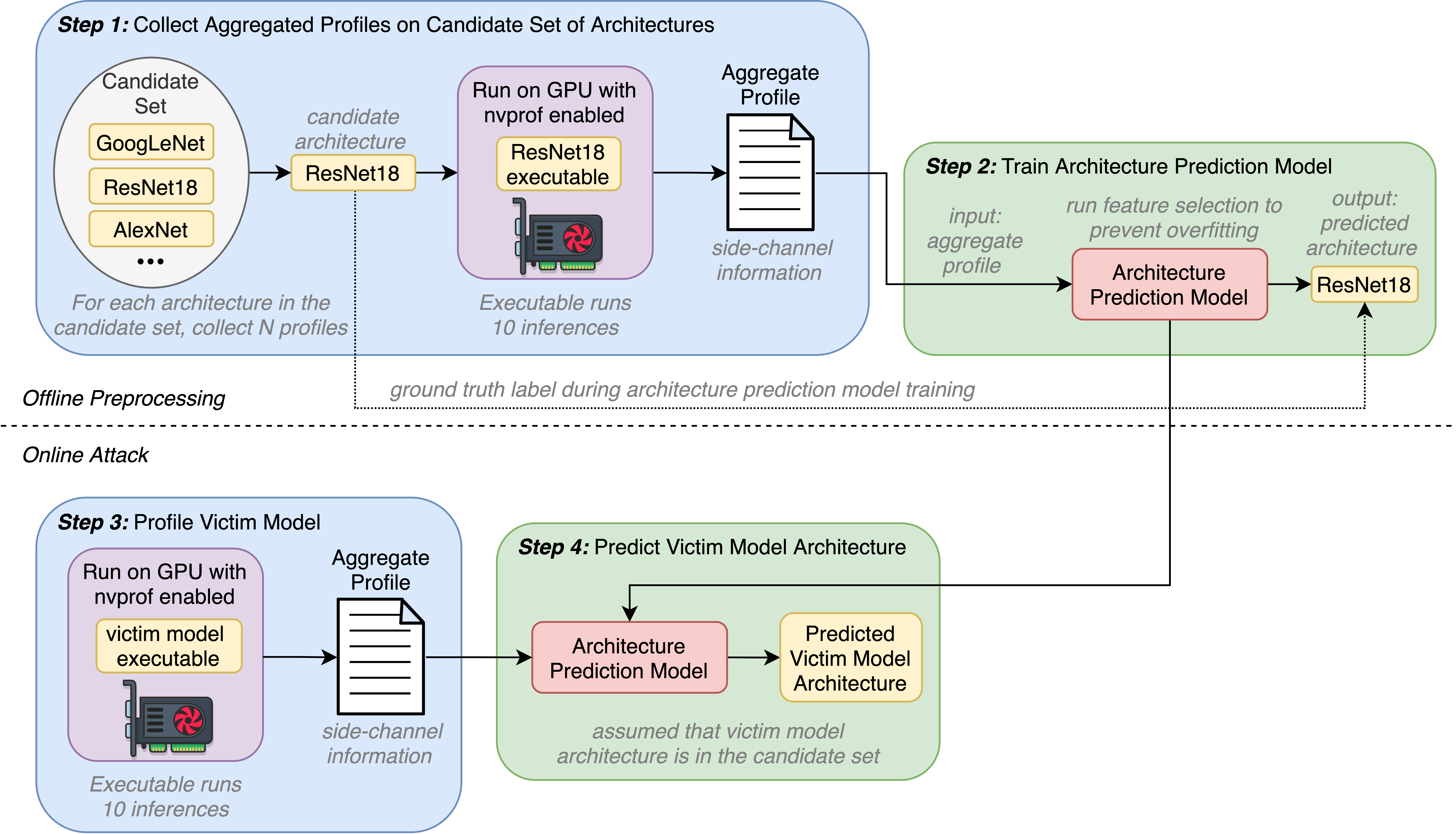}
\caption{InferNet Attack Overview. InferNet involves an \textit{offline preprocessing} phase that includes collecting GPU profiles, cleaning the dataset, and training the architecture prediction model, followed by an \textit{online attack} phase. For instance, ResNet18 model is selected from the candidate set $C$ and executed on a GPU for 10 inferences to collect its aggregate GPU profile $\mathcal{H}$. The architecture prediction model $\mathcal{A}$ uses \texttt{ResNet18} architecture name as the ground truth to learn the correlations between GPU profile and the architecture.}
\label{fig:InferNet}
\end{figure*}

Using the aforementioned insights, we present InferNet: an attack paradigm to extract the DNN architectures using aggregate GPU profiles as a side-channel (illustrated in ~\autoref{fig:InferNet}).

\subsection{Threat Model}

The adversary aims to conduct an accurate and cost-effective model extraction attack using a coarse-grained and non-intrusive side-channel. We assume that the adversary has black-box access to standard DNN architectures, including state-of-the-art vision and large language models, through an executable file that they can run on a GPU-enabled system. They have no prior knowledge of the target DNN architecture and its parameters, and may only query the model by providing an input $x$ and observing the output $y = f^\theta(x)$.

The adversary is assumed to have non-root access to the GPU-enabled system running the target model and access to profiling tools such as \texttt{nvprof}. While NVIDIA introduced an administrator-only access flag in later versions of {\nvprof} to address security concerns~\cite{rendered_insecure}, this restriction is not uniformly enforced across cloud platforms. Cloud environments may deploy different versions of the CUDA Toolkit---\eg CUDA 10.2 uses {\nvprof} 10.2, CUDA 11.0 uses {\nvprof} 11.0, and CUDA 11.1 onward migrates to Nsight Systems---with the specific profiler version depending on the provider's deployment choices. In some deployments, these environments allow profiling by default for non-root users. Additionally, even when newer profilers enforce access restrictions, legacy CUDA toolchains may remain usable in containerized user-space environments~\cite{leaky_dnn}. Accordingly, InferNet applies to deployments in which aggregate GPU profiling remains available to the attacker due to platform configuration, legacy software stacks, or equivalent profiling exposure; we do not claim that this access model is universal across all MLaaS platforms.

We assume that the adversary does not tamper with system assets: software or hardware. The attack does not require access to the PCIe bus, operating system (OS), or any internal system components. This makes our attack non-intrusive, which relies solely on profiling data generated during the execution of the target model. \newline

\noindent
\textbf{Why aggregate profiles?}\quad
Aggregate profiles are less sensitive to instrumentation differences across profiler versions. Aggregation suppresses noise arising from non-deterministic GPU scheduling across multiple processes and kernels---GPU's internal scheduling policies may vary between runs, affecting timing and execution order. Aggregate profiles also mitigate variability introduced by thermal throttling and power management, where dynamic changes in performance states (P-states) alter runtime characteristics across inferences. Additionally, aggregate metrics average out fluctuations caused by internal resource contention, such as shared memory and SM-level warp scheduling. These effects introduce instability in fine-grained (time series) profiles, whereas aggregate data provide a more stable and representative fingerprint. Moreover, our feature selection process further reduces the influence of residual noise by retaining only the most predictive components of the profile, enabling high inference accuracy (see \autoref{sec:eval}) with minimal complexity.

\subsection{Architecture Extraction Attack}
We formulate a model extraction attack where an adversary infers the architecture of a \textit{victim model} $f_v^\theta$ with architecture $f_v$ using aggregate GPU profiles $\mathcal H(f_v^\theta)$ as a side-channel. Our attack methodology includes an \textit{offline preprocessing} phase that includes collecting GPU profiles, cleaning the dataset, and training the architecture prediction model, followed by an \textit{online attack} phase (see \autoref{fig:InferNet}). \newline

\noindent
\textbf{Experiment Setup.}\quad
We take a diverse set of 90 candidate DNN architectures, including widely used LLMs (\eg GPT~\cite{brown2020language}, BERT~\cite{devlin2018bert}), vision models (\eg ResNet~\cite{he2016deep}, GoogLeNet~\cite{szegedy2015going}), and vision transformer models (\eg MaxViT~\cite{tu2022maxvit}, ViT~\cite{dosovitskiy2020image}). The candidate models are \textit{pretrained} and sourced from two major AI/ML frameworks: TensorFlow (v2.17.1)~\cite{abadi2015tensorflow}, and PyTorch (v2.3.0)~\cite{paszke2019pytorch}, where 24 models are implemented in both of the frameworks. We import TensorFlow vision models from \texttt{tf.keras.applications}, and LLMs from HuggingFace's AutoModels~\cite{huggingface}. Similarly, for PyTorch, we import models from TorchVision (v0.18.0) and TorchText (v0.18.0). The complete list of the models used in this paper is provided in \autoref{app:models}.

To further enhance the robustness of our approach, we conduct our experiments on two different NVIDIA GPUs: \texttt{Tesla T4}~\cite{nvidia_tesla_t4} on Google Colaboratory~\cite{colab}, and \texttt{Quadro RTX 8000}~\cite{nvidia_quadro_rtx} on a local server (running Ubuntu 24.04). By using two separate experiment setups, we demonstrate that our approach is generalizable to different hardware and software platforms.

A complete summary of the experimental settings used for architecture prediction, including profile collection, feature engineering, classifier configuration, and evaluation metrics, is provided in \autoref{app:hyper}.\newline

\noindent
\textbf{Collecting Aggregate Profiles.}\quad
We define a candidate architectures set $C = \{f_{c_1}, f_{c_2}, ... f_{c_n}\}$ comprising popular large language models (LLMs) and vision models. For each candidate $f_{c_i} \in C$, we collect $N = 20$ aggregate profiles $\mathcal H(f_{c_i}^{\theta_i})$ using \texttt{nvprof} in \textit{aggregate} mode to construct a data sample $S_i = \{\mathcal H(f_{c_i}^{\theta_i}), f_{c_i}\}^N$. Every profile contains combined metrics, such as kernel runtimes, memory operations, and system-level resource utilization, for ten (10) inferences with \texttt{zero} input. We avoid batch execution to reflect realistic attack conditions where the batch sizes are unknown to the adversary.

We consolidate all the profiles from the candidate architectures into a comprehensive training dataset $D = \bigcup_i S_i$. With 90 candidates and 2 GPU platforms, our dataset accumulates $|D| = 90 \times 2 \times 20 = 3600$ GPU profiles representing 36,000 inference runs. This dataset represents the computational characteristics of the architectures in the candidate set $C$, where each kernel name serves as a distinct \textit{feature} for the architecture's prediction model. \newline

\noindent
\textbf{Data Cleaning.}\quad
In the combined dataset $D$,  some GPU kernels and memory operations appear only for specific architectures; therefore, creating a sparse dataset. We address this by adding binary \textit{indicator} columns for each kernel that denote the kernel's presence or absence in a given profile. Then, we fill any missing values with the mean of the corresponding feature across the dataset. Finally, we account for the different magnitudes of metrics by normalizing and scaling all the features in the dataset.

We perform \textit{feature selection} to improve the signal-to-noise ratio and reduce the risk of overfitting. In this regard, we use Recursive Feature Elimination (RFE) to identify and retain the most relevant features. The resulting trained prediction model creates a more reliable mapping between aggregate profiles and the target architecture $f_{v}$. \newline

\noindent
\textbf{Training an Architecture Prediction Model.}\quad
We train an architecture prediction model $\mathcal{A}$ on the clean dataset $D$ to learn the mapping between aggregate GPU profiles $\mathcal{H}$ and their corresponding architectures $f_{c_i}$. Formally, $\mathcal{A}$ attempts to solve the problem:
\begin{equation} \label{eq:a}
    \mathcal{A}( \mathcal{H}(f_{v}^{\theta})) = \mathbb{E} [ f_{v} | \mathcal{H}(f_{v}^{\theta}) ] = \hat{f}_{v}
\end{equation}
where $\hat{f}_v$ is the predicted architecture of the victim model, and $\mathcal A$ is typically a machine learning model. The training process uses simple supervised classifier models, \ie Logistic Regression, Neural Network, KNN, Nearest Centroid, Naive Bayes, Random Forest, and AdaBoost. Using multiple models allows us to optimize the attack for different architectures.\newline

\noindent
\textbf{Online Attack.}\quad
The online attack phase focuses on extracting the architecture of an unknown victim model. We use a binary executable to run the victim model $f_v^\theta$ on a GPU-enabled system, and use \texttt{nvprof} to generate an \textit{aggregate} profile $\mathcal{H}(f_v^\theta)$. This profile is fed into the trained prediction model $\mathcal{A}$, which outputs a predicted architecture $\hat{f}_{v} = \mathcal{A}( \mathcal{H}(f_{v}^{\theta}))$.
\section{InferNet Attack Evaluation}
\label{sec:eval}

In this section, we assess the accuracy of InferNet across various attack scenarios. We begin by evaluating a practical attack scenario using vanilla settings. Next, we optimize the attack, enhancing its resilience across diverse use cases. Finally, we demonstrate the scalability of the attack with respect to future DNN architectures. For model training, unless otherwise stated, we used the default hyperparameters of the corresponding scikit-learn implementation, with the exceptions listed in \autoref{app:hyper}.

\subsection{Practical Attack}
\label{sec:base_attack}
We begin by evaluating whether the aggregate GPU profiles can reveal DNN architectures in a \textit{practical} scenario with standard settings. In this setting, the victim model $f_v^\theta$ and candidate model $f_{c_i}^{\theta_i}$ have the same number of output classes. The profiles are split into two datasets: $D_1$ for training (75\%) and $D_2$ for testing (25\%). Dataset $D_1$ represents profiles collected during the \textit{offline preprocessing} phase of the attack, while $D_2$ simulates victim profiles collected for the \textit{online attack}. We stratify the data by the DNN architecture names to ensure that each DNN architecture is represented in both the training and test datasets. We train seven (7) architecture prediction models $\mathcal{A}$ on $D_1$ and evaluate their performance on $D_2$. Each model is trained using different subsets of profile features (\eg GPU kernel features, API call features) as shown in \autoref{table:arch_pred_acc_by_data_subset}. \newline

\noindent
\textbf{Key Findings.}\quad
Our key findings are (refer to \autoref{table:arch_pred_acc_by_data_subset}):

\noindent
\textit{Effectiveness of architecture prediction models:}
The results confirm that aggregated profiles can effectively reveal victim architectures in a common scenario. Most models, including Neural Network, K Nearest Neighbors, Naive Bayes, and Random Forest, achieve near-perfect accuracy ($\ge 96\%$) on $D_2$ when trained on all 849 features. Models like Random Forest, Naive Bayes, and K Nearest Neighbors exhibit robustness even when trained on reduced feature sets.

\noindent
\textit{Impact of feature selection:}
GPU kernel features are the most critical for architecture prediction. Models trained only on GPU kernel features achieve accuracy comparable to those trained on all features. API call features and system-level metrics contribute less, as training exclusively on these features results in significant accuracy drops.

\noindent
\textit{Overfitting on system features:}
We observe spurious correlations in system-level metrics that do not generalize across datasets. Models trained only on the system features overfit on the training data $D_1$. For example, K Nearest Neighbors achieves 100\% accuracy on 
$D_1$ but drops to 47.3\% on $D_2$. Similarly, the accuracy for Random Forest, the best model overall, drops from 100\% to 61.3\%. These results suggest that system-level metrics, may reflect specific runtime conditions, hardware configurations, or workload distributions that are unique to the input data and not the DNN architecture.

\noindent
\textit{Impact of memory management kernels:}
Memory-related GPU kernels, such as \texttt{CUDA\_memcpy\_HtoD}, are less informative for architecture prediction. Removing memory management features from the GPU kernel subset causes minimal ($\le 1\%$) accuracy loss for all models except AdaBoost. This highlights that memory operations contribute little to distinguishing architectures.

\begin{table*}[t]
\caption{Top1 Train ($D_1$) / Test ($D_2$) accuracy of architecture prediction models $\mathcal A$ by the subset of profile features used to train $\mathcal A$. System features, GPU kernels, and API calls correspond to the system signals, GPU kernels, and API calls from Table~\ref{tab:profile} and Table~\ref{tab:system_signals}, respectively; whereas, Indicator features denote the presence or absence of a profile feature. Train and test datasets $D_1$ and $D_2$ contain profiling data for all 56 TorchVision and 4 TorchText models.}
\centering
\begin{tabular}{@{}llllllll@{}}
\toprule
\multirow{2}{*}{\begin{tabular}[c]{@{}l@{}}Architecture \\ Prediction \\ Model $\mathcal A$\end{tabular}} & \multicolumn{7}{c}{\textbf{Subset of Profile Features Used to Train Architecture Prediction Model}}\\ \cmidrule(l){2-8} 
& \textbf{\begin{tabular}[c]{@{}l@{}}All\\ (849)\end{tabular}} & \textbf{\begin{tabular}[c]{@{}l@{}}System\\ (27)\end{tabular}} & \textbf{\begin{tabular}[c]{@{}l@{}}No System\\ (822)\end{tabular}} & \textbf{\begin{tabular}[c]{@{}l@{}}GPU Kernel \\ (447)\end{tabular}} & \textbf{\begin{tabular}[c]{@{}l@{}}API Calls\\ (375)\end{tabular}} & \textbf{\begin{tabular}[c]{@{}l@{}}Indicator\\ (68)\end{tabular}} & \textbf{\begin{tabular}[c]{@{}l@{}}No Indicator\\ (781)\end{tabular}} \\ \midrule
Logistic Regression                                                                               & 99.9/99.1                                                    & 36.3/31.5                                                      & 99.9/99.2                                                          & 99.3/99.4                                                            & 94.2/83.4                                                          & 57.4/52.8                                                         & 99.9/99.0                                                             \\
Neural Network                                                                                    & 100/99.1                                                     & 65.9/58.7                                                      & 100/98.7                                                           & 100/99.9                                                             & 99.4/84.3                                                          & 57.1/52.3                                                         & 100/98.8                                                              \\
K Nearest Neighbors                                                                               & 100/96.0                                                     & 100/47.3                                                       & 100/95.8                                                           & 100/98.8                                                             & 100/66.4                                                           & 56.2/55.5                                                         & 100/95.3                                                              \\
Nearest Centroid                                                                                  & 99.3/97.8                                                    & 34.6/32.8                                                      & 99.1/97.7                                                          & 98.4/98.5                                                            & 81.3/71.2                                                          & 55.5/57.4                                                         & 99.0/97.8                                                             \\
Naive Bayes                                                                                       & 100/99.7                                                     & 47.4/43.6                                                      & 100/99.9                                                           & 100/99.8                                                             & 100/99.6                                                           & 57.2/52.0                                                         & 100/99.7                                                              \\
Random Forest                                                                                     & 100/100                                                      & 100/61.3                                                       & 100/100                                                            & 100/100                                                              & 100/100                                                            & 56.7/52.4                                                         & 100/100                                                               \\
AdaBoost                                                                                          & 42.5/41.8                                                    & 6.11/4.02                                                      & 45.8/44.9                                                          & 46.7/48.6                                                            & 24.7/20.8                                                          & 15.4/12.3                                                         & 37.5/36.6                                                             \\ \bottomrule
\end{tabular}
\label{table:arch_pred_acc_by_data_subset}
\end{table*}

\subsection{Attack Optimization for Sparse Data}
\label{sec:data_optimization}

Since not all features strongly indicate the DNN architecture, we investigate how small a subset of features is sufficient for high prediction accuracy. We focus primarily on the 450 \textit{non-memory} GPU kernel features (see \textit{Finding 2} in \S~\ref{sec:base_attack}). Using Recursive Feature Elimination (RFE)~\cite{rfe}, we rank the kernel features based on their importance and train the architecture prediction models $\mathcal{A}$ on $D_1$ from the baseline attack (\autoref{sec:base_attack}) using only the top features.  \newline

\begin{figure}[t]
\centering
\includegraphics[width=0.65\linewidth]{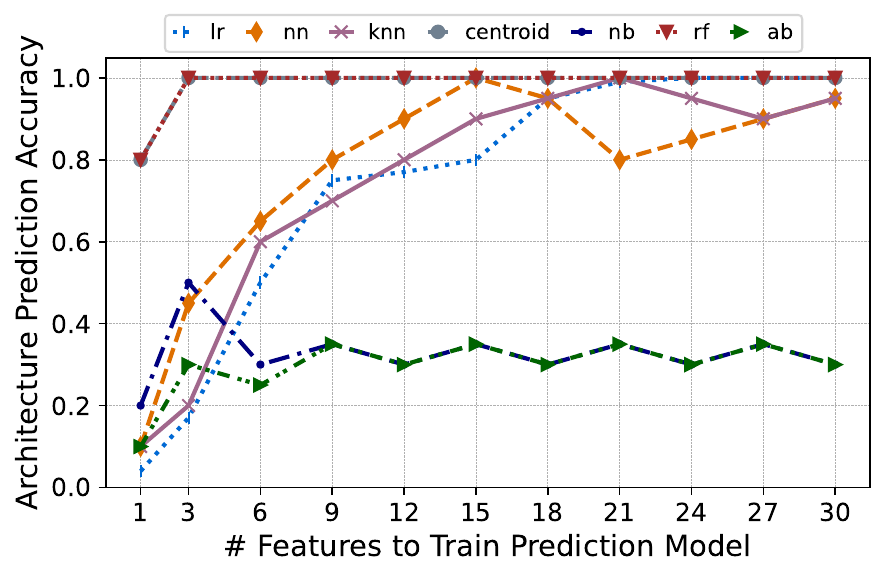}
\caption{Top1 accuracy of architecture prediction models $\mathcal A$ on profiles in $D_2$ by number of GPU kernel features used to train $\mathcal A$.}
\label{fig:arch_pred_acc_by_num_features}
\end{figure}

\noindent
\textbf{Key Findings.}\quad
Our results, illustrated in \autoref{fig:arch_pred_acc_by_num_features}, validate the hypothesis that a small number of features can achieve comparable accuracy to models trained on the full feature set. These findings highlight the importance of feature selection in choosing a minimal feature subset, mitigating overfitting, and reducing computational overhead.

\noindent
\textit{The top features are highly effective:}
Models like K Nearest Neighbors, Nearest Centroid, Naive Bayes, and Random Forest achieve the same high accuracy (\(\sim \)100\%) using only the top 3 features as they do with over 450 features. The top three features, shown in \autoref{fig:top_features}, include \texttt{Time (ms) volta\_sgemm\_128x64\_tn}, \texttt{Avg (us) volta\_sgemm\_128x64\_tn}, and \texttt{Time (ms) void splitKreduce\_kernel<int=A, int=B, ...>}\footnote{Truncated long kernel name.}.

\noindent
\textit{Feature requirements vary by prediction model $\mathcal{A}$:}
Neural Networks and Logistic Regression require more features, \ie 17 and 22, respectively, to achieve peak accuracy. This suggests that not all models capture the same activity patterns and therefore prioritize different feature sets for predictions.

The top-kernel features are informative because the paper’s mapping analysis ties kernels to layer structure and variant-specific execution behavior; therefore, feature selection suppresses environment-specific noise rather than discovering arbitrary correlations.

\subsection{Attack Resilience Against Modified Architectures}
\label{sec:advance_attack}
We extend our attack to a more challenging scenario where the victim model $f_v^\theta$ is modified by additional final layer(s) with a different number of output classes than the original base model. We simulate this scenario by taking a subset of $D_1\prime$ from ~\autoref{sec:base_attack}, containing only the 56 TorchVision models. The $D_1\prime$ dataset represents candidate models \textit{pretrained} on ImageNet~\cite{imagenet} with 1000 output classes. However, we construct a new dataset $D_3$ with vision models trained in CIFAR10~\cite{cifar10} with only 10 output classes.   \newline

\noindent
\textbf{Key Findings.}\quad
For evaluation, we train the prediction models $\mathcal{A}$ on $D_1\prime$, and test them on $D_3$. The features used for training include the top 3, top 25, and all 450 non-memory GPU kernel features identified in \S~\ref{sec:data_optimization}. Our findings, from \autoref{table:arch_pred_acc_test}, are as follows:

\noindent
\textit{High correlation with top features:} The top 3 features: (i) \texttt{Time (ms) volta\_sgemm\_128x64\_tn}, (ii) \texttt{Avg (us) volta\_sgemm \_128x64\_tn}, and (iii) \texttt{Time (ms) void splitKreduce\_kernel<int=A, int=B, ...>}\footnote{Truncated long kernel name.}, continue to show strong predictive power. This
suggests that these kernel features are highly correlated with the architectural characteristics of DNNs and are robust to variations in the number of output classes.

\noindent
\textit{Performance gain with top 25 features:}
Expanding the feature set to the top 25 features improves the performance of Logistic Regression and Neural Network models, with accuracy on the test set $D_3$ of 100\%, and 95.2\%, respectively. Therefore, with additional features available, a broader range of models can be used for architecture prediction with high confidence.

\noindent
\textit{Overfitting with all features:}
Using all 450 non-memory kernel features introduces overfitting for models that prioritize memory as well as non-memory kernels. In particular, the accuracies for Logistic Regression and Neural Networks drop to 58.7\% and 61.4\%, respectively, on $D_3$. This performance is far lower than their predictions using all GPU kernels (see \autoref{table:arch_pred_acc_by_data_subset}), with accuracies of 99.4\%, and 99.9\%, respectively.  Random Forest, however, remains robust and consistently achieves 100\% accuracy.

\begin{table}[t]
\caption{Top1 Train ($D_1\prime$) / Test ($D_3$) architecture prediction accuracy for architecture prediction models $\mathcal A$ by the top 3, 25, and 450 GPU kernel features (without memory data) used to train $\mathcal A$. $D_1\prime$ consists of profiles generated by models trained on ImageNet, and $D_2$ consists of profiles generated by models trained on CIFAR10.}
\centering
\begin{tabular}{@{}llll@{}}
\toprule
\begin{tabular}[c]{@{}l@{}}\textbf{Architecture}  \\
\textbf{Prediction Model }$\mathcal A$\end{tabular} & \textbf{\begin{tabular}[c]{@{}l@{}}Top 3\\ Features\end{tabular}} & \textbf{\begin{tabular}[c]{@{}l@{}}Top 25\\ Features\end{tabular}} & \textbf{\begin{tabular}[c]{@{}l@{}}All 450\\ Features\end{tabular}} \\
\midrule
Logistic Regression                                                                    & 38.2/40.5                                                         & 99.9/100                                                           & 97.6/58.7                                                           \\
Neural Network                                                                         & 20.3/18.9                                                         & 100/95.2                                                           & 100/61.4                                                            \\
K Nearest Neighbors                                                                    & 100/100                                                           & 100/100                                                            & 100/85.9                                                            \\
Nearest Centroid                                                                       & 100/100                                                           & 100/100                                                            & 96.8/81.4                                                           \\
Naive Bayes                                                                            & 98.7/94.3                                                          & 100/100                                                            & 100/63.5                                                            \\
Random Forest                                                                          & 100/100                                                           & 100/100                                                            & 100/100                                                             \\
AdaBoost                                                                               & 15.4/17.6                                                         & 37.9/35.4                                                          & 28.6/26.3                                                           \\
\bottomrule
\end{tabular}
\label{table:arch_pred_acc_test}
\end{table}

The results confirm that our attack method remains effective even when the DNN is fine-tuned or modified. This demonstrates that GPU kernel features serve as a reliable side channel for extracting DNN architectures, regardless of changes to the model architecture.

The high predictive value of the top-ranked features is consistent with the kernel-to-layer mapping discussed in Section~\ref{sec:insights}: these kernels are not arbitrary correlates, but compact summaries of architecture-dependent operations. At the same time, our results show that indiscriminate inclusion of many low-value features can reduce transfer performance, which is why we treat full-feature results as an ablation rather than as a primary attack configuration.


\subsection{Scaling Attack to Future Architectures}

\begin{figure}[t]
\centering
\includegraphics[width=0.65\linewidth]{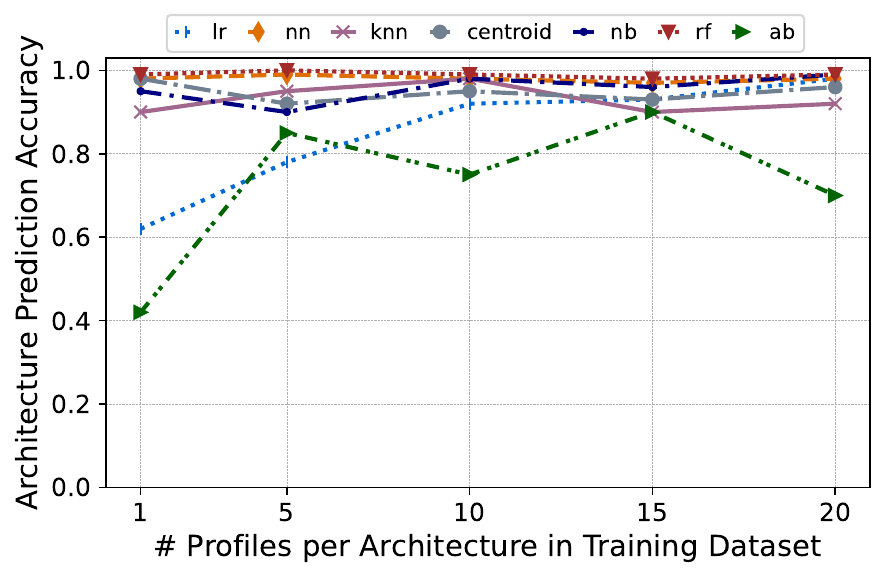}
\caption{Top1 accuracy of architecture prediction models $\mathcal A$ on profiles in dataset $D_3$ by number of profiles per candidate architecture from $D_1$ used to train $\mathcal A$.}
\label{fig:arch_pred_acc_by_dataset_size}
\end{figure}

Adding new candidate DNN architectures to the attack's scope requires additional profiles and significant computation for retraining the prediction model $\mathcal{A}$. For reference, generating a single profile for DNN execution takes approximately 45-120 seconds, and generating the complete dataset $D_1$ requires 20+ hours. In contrast, the \textit{online attack} phase is computationally inexpensive, with less than one minute needed to profile the victim model and predict its architecture. Reducing the time spent in the \textit{offline preprocessing} phase is critical for enhancing the practicality of the attack.

We analyze the computational efficiency of the InferNet attack by reducing the training dataset size -- and therefore, the time for collecting profiles -- without compromising the accuracy.
We hypothesize that fewer profiles per architecture are sufficient for training accurate architecture prediction models. To test this, we train the prediction models $\mathcal{A}$ on subsets of $D_1\prime$ with a varying number of profiles per candidate architecture. We limit the training to the top 25 GPU kernel features identified in \autoref{sec:advance_attack}. Finally, we test $\mathcal{A}$ on $D_3$ (profiles of victim models trained on CIFAR10) to evaluate the impact of dataset size reduction.  \newline

\noindent
\textbf{Key Findings.}\quad
\autoref{fig:arch_pred_acc_by_dataset_size} confirms that architecture prediction models can generalize effectively with minimal training data.  For all models except Logistic Regression and AdaBoost, a single profile per candidate architecture is sufficient to achieve the same accuracy as training on the full dataset of 20 profiles. Similarly, these models show very low variance for the top 25 features which indicates that repeated executions of DNN models often result in identical GPU profiles. 

Reducing the training data requirements offers significant practical advantages. Collecting just one profile per architecture reduces the \textit{offline preprocessing} phase to under 30 minutes (\ie 97.5\% time reduction). Similarly, adding a new architecture to the candidate set requires only one additional profile or just over one minute to fine-tune the attack. These findings highlight the scalability of InferNet and its adaptability to expanding candidate model pools.
\section{Generalizability Evaluation of InferNet}
\label{sec:analysis}
While our evaluation results in Section~\ref{sec:eval} indicate that InferNet is reliably architecture-agnostic; it is strongest when the attacker can match or approximate the victim’s software and hardware stack. This section evaluates the generalizability of InferNet across various deployment scenarios, including pruned DNNs, different GPU hardware, and multiple AI/ML frameworks. 

\subsection{Evaluation Over Pruned Models}
DNNs are often pruned to optimize inference performance and reduce computational overhead. To evaluate whether InferNet is effective against such models, we create a dataset $D_p$ using profiles of 56 TorchVision models trained on CIFAR10~\cite{cifar10} and pruned to 50\% capacity. Architecture prediction models $\mathcal{A}$ are trained on the top 25 non-memory GPU kernel features from the unpruned dataset $D_1$ (from \S~\ref{sec:base_attack}) and tested on $D_p$. This setup also simulates the scenario where the DNN architecture is modified and fine-tuned with a different number of output classes (see \S\ref{sec:advance_attack}). \newline

\noindent
\textbf{Key Findings.}\quad
The results, as shown in \autoref{table:prune_acc}, demonstrate that InferNet generalizes well to pruned models with minimal performance degradation. Random Forest and Naive Bayes models achieve 100\% architecture prediction accuracy. However, the rest of the models show a slight performance decline from the non-pruned test set (see \autoref{table:arch_pred_acc_test}). Performance for some of the models, like Logistic Regression and Neural Network, improves significantly in Top5 evaluation, indicating that the true architecture is often among the top five predictions. We also include DNN Family prediction to analyze if the prediction models $\mathcal{A}$ recognize the DNN family, if they fail to recognize the architecture itself. Most models, except AdaBoost, deliver reliable performance, with many achieving 100\% accuracy in predicting the DNN family.

\begin{table}[t]
\small
\caption{Top1/Top5 test accuracy of architecture prediction models trained on profiles from unpruned (ImageNet) models and tested on profiles of pruned (CIFAR10) models.}
\centering
\begin{tabular}{@{}l|ccc@{}}
\toprule
\textbf{Architecture} & \multirow{2}{*}{\textbf{Top1 \%}} & \multirow{2}{*}{\textbf{Top5 \%}} & \multirow{2}{*}{\textbf{Family \%}} \\
\textbf{Prediction Model $\mathcal{A}$} &  &  & \\
\midrule
\textbf{Logistic Regression} & 94.2 & 99.7 & 100 \\
\textbf{Neural Network} & 81.3 & 90.9 & 87.5 \\
\textbf{KNN} & 96.8 & 97.5 & 100 \\
\textbf{Nearest Centroid} & 97.5 & 96.9 & 100 \\
\textbf{Naive Bayes} & 100 & 100 & 100 \\
\textbf{Random Forest} & 100 & 99.9 & 100 \\
\textbf{AdaBoost} & 30.4 & 41.3 & 46.1 \\
\bottomrule
\end{tabular}
\label{table:prune_acc}
\end{table}

\subsection{Evaluation Across GPUs}
In practice, a diverse range of GPUs is available for training and running DNN models. Therefore, it is crucial to establish the effectiveness of the attack when the adversary does not have easy access to the same GPU for both offline and online phases.

We collect two datasets $D_2\prime$ and $D_3\prime$ using profiles \texttt{NVIDIA Quadro RTX 8000} GPU; whereby, $D_2\prime$ contains profiles of candidate models trained on ImageNet~\cite{imagenet} and $D_3\prime$ contains profiles of models trained on CIFAR10~\cite{cifar10}. Additionally, we re-use the datasets $D_1\prime$, $D_2$, and $D_3$ collected from \texttt{NVIDIA Tesla T4} GPU in \S~\ref{sec:base_attack} and \S~\ref{sec:advance_attack}. Both datasets use the top 25 non-memory GPU kernel features for training and testing. For cross evaluation, training data consists of profiles from one GPU, while the test data comes from the other GPU, and vice versa. \newline

\begin{table}[t]
\caption{Top1/Top5 and Architecture Family test accuracy of architecture prediction models trained on profiles from one GPU and tested on profiles from another GPU. Training data consists of profiles of candidate models trained on ImageNet, while test data consists of profiles of candidate models trained on CIFAR10.}
\centering
\begin{tabular}{@{}lllll@{}}
\toprule
\multirow{2}{*}{\begin{tabular}[c]{@{}l@{}}\textbf{Architecture}\\ \textbf{Prediction Model} $\mathcal A$\end{tabular}} & \multicolumn{2}{c}{\textbf{\begin{tabular}[c]{@{}c@{}}Quadro\textsubscript{train} \\Tesla\textsubscript{test}\end{tabular}}} & \multicolumn{2}{c}{\textbf{\begin{tabular}[c]{@{}c@{}}Tesla\textsubscript{train} \\ Quadro\textsubscript{test}\end{tabular}}} \\ \cmidrule(l){2-5} 
& \textit{Top1/Top5}                        & \begin{tabular}[c]{@{}l@{}}\textit{Fam.}\end{tabular}                       & \textit{Top1/Top5}                        & \begin{tabular}[c]{@{}l@{}}\textit{Fam.}\end{tabular}                       \\ \midrule
\textbf{Logistic Regression}                                                                                                       & 34.1/74.2                        & 78.5                                                                        & 28.5/56.8                        & 48.1                                                                        \\
\textbf{Neural Network}                                                                                               & 10.7/28.3                        & 32.5                                                                        & 12.0/25.5                        & 31.1                                                                        \\
\textbf{KNN}                                                                                                      & 35.2/42.5                        & 74.1                                                                        & 28.1/26.8                        & 51.2                                                                        \\
\textbf{Nearest Centroid}                                                                                                 & 35.6/42.0                        & 76.3                                                                        & 28.3/28.5                        & 50.5                                                                        \\
\textbf{Naive Bayes}                                                                                              & 71.4/73.5                        & 96.7                                                                        & 74.6/74.1                        & 96.3                                                                        \\
\textbf{Random Forest}                                                                                              & 42.5/74.1                        & 81.4                                                                        & 45.3/78.6                        & 59.7                                                                        \\
\textbf{AdaBoost}                                                                                                 & 6.2/28.5                        & 26.1                                                                        & 34.1/62.8                        & 65.2                                                                        \\ \bottomrule
\end{tabular}
\label{table:cross_gpu}
\end{table}

\noindent
\textbf{Key Findings.}\quad
The results from \autoref{table:cross_gpu} show that InferNet achieves varying levels of accuracy across GPUs. For example, Logistic Regression achieves only 34.1\% Top1 accuracy when trained on Quadro and tested on Tesla, and even lower accuracy when reversed. Neural Network accuracy is similarly low, at 10.7\% and 12\% Top1 accuracy in both cross-GPU scenarios, respectively. However, models like Naive Bayes and Random Forest demonstrate stronger cross-GPU transfer across the two evaluated NVIDIA GPUs, with Naive Bayes achieving 71.4\% Top1 accuracy when trained on Quadro and tested on Tesla, and 74.6\% when trained on Tesla and tested on Quadro. Interestingly, even when architecture prediction accuracy drops, family prediction accuracy remains relatively high for most models.

Our hardware selection of Tesla T4 and Quadro RTX 8000 represents two widely used deployment tiers within NVIDIA’s Turing generation: a data-center accelerator and a high-end workstation GPU. The evaluation reinforces that InferNet exploits fundamental architectural behaviors rather than specific hardware instance noise. Successfully transferring the attack between two vastly different operational tiers of the Turing architecture, \ie the T4 and the RTX 8000, indicates that the aggregate profile acts as a robust, hardware-agnostic side-channel within an architectural generation.

Analysis of the combined data shows that the GPU kernels used by DNN architectures are largely consistent across GPUs. This consistency enables some generalizability. However, the observed performance drop highlights the need for further tuning or additional training data to improve the cross-GPU transferability of InferNet.

\subsection{Evaluation Across ML Frameworks}

Given the effectiveness of InferNet in various scenarios, the next important question is whether the adversary must know the framework used to implement the victim model or if architecture prediction models can be generalized across frameworks. To evaluate this, we create a dataset of GPU aggregate profiles for 24 vision DNNs implemented in TensorFlow v2.17.1 (see \autoref{app:models}). We choose those DNNs that are also available in PyTorch, and they are part of our $D_1\prime$ profiles dataset. To collect 20 profiles each for the TensorFlow models to construct a dataset $D_t$. For evaluation, we train using profiles from one framework and test with profiles from the other framework, and vice versa.
The results indicate that TensorFlow DNNs utilize a largely \textit{disjoint} set of GPU kernels compared to PyTorch. This results in a significant accuracy drop when using PyTorch-trained models to predict TensorFlow architectures (see \autoref{table:cross_framework}).

\begin{table}[t]
\caption{Top1/Top5 and Family accuracy of architecture prediction models trained on profiles from one ML framework (PyTorch) and tested on profiles from another (TensorFlow), and vice versa.}
\centering
\begin{tabular}{@{}lcccc@{}}
\toprule
\multirow{2}{*}{\begin{tabular}[c]{@{}l@{}}\textbf{Architecture}\\ \textbf{Prediction Model} $\mathcal A$\end{tabular}} & \multicolumn{2}{c}{\textbf{\begin{tabular}[c]{@{}c@{}}PyTorch\textsubscript{train} \\ TensorFlow\textsubscript{test}\end{tabular}}} & \multicolumn{2}{c}{\textbf{\begin{tabular}[c]{@{}c@{}}TensorFlow\textsubscript{train} \\ PyTorch\textsubscript{test}\end{tabular}}} \\
\cmidrule(l){2-5} 
& \textit{Top1/Top5} & \begin{tabular}[c]{@{}l@{}}\textit{Fam.}\end{tabular} & \textit{Top1/Top5} & \begin{tabular}[c]{@{}l@{}}\textit{Fam.}\end{tabular} \\
\midrule
\textbf{Logistic Regression} & 26.1/50.3 & 67.4 & 20.8/51.7 & 41.3 \\
\textbf{Neural Network} & 9.3/22.8 & 27.6 & 11.2/22.1 & 29.4 \\
\textbf{KNN} & 43.5/52.2 & 82.7 & 36.7/36.3 & 59.9 \\
\textbf{Nearest Centroid} & 32.8/41.7 & 70.1 & 26.8/32.4 & 48.9 \\
\textbf{Naive Bayes} & 58.2/60.3 & 80.0 & 64.1/65.4 & 78.3 \\
\textbf{Random Forest} & 51.6/63.2 & 91.0 & 41.1/54.3 & 66.9 \\
\textbf{AdaBoost} & 14.1/27.3 & 32.6 & 17.9/29.5 & 51.7 \\
\bottomrule
\end{tabular}
\label{table:cross_framework}
\end{table}

\begin{table}[t]
\caption{Top1/Top5 test accuracy of architecture prediction models trained on combined PyTorch and TensorFlow profiles.}
\centering
\begin{tabular}{@{}l|ccc@{}}
\toprule
\textbf{Architecture} & \multirow{2}{*}{\textbf{Top1 \%}} & \multirow{2}{*}{\textbf{Top5 \%}} & \multirow{2}{*}{\textbf{Family \%}} \\
\textbf{Prediction Model $\mathcal{A}$} &  &  & \\
\midrule
\textbf{Logistic Regression} & 93.1 & 98.6 & 100 \\
\textbf{Neural Network} & 84.4 & 91.7 & 92.3 \\
\textbf{KNN} & 96.5 & 96.9 & 100 \\
\textbf{Nearest Centroid} & 95.7 & 95.8 & 100 \\
\textbf{Naive Bayes} & 100 & 100 & 100 \\
\textbf{Random Forest} & 100 & 100 & 100 \\
\textbf{AdaBoost} & 32.8 & 43.2 & 48.7 \\
\bottomrule
\end{tabular}
\label{table:heterogeneous}
\end{table}

In cross-framework evaluations (PyTorch to TensorFlow), while raw Top-1 accuracy drops, the security implication remains severe. The ability to identify the model family (e.g., VGG) reduces the architectural search space for an adversary from thousands of possibilities to a handful of known variants. This facilitates targeted model-stealing, as the adversary can focus their resources on a narrow set of probable architectures once the family is confirmed.\newline

\noindent
\textbf{Optimization.}\quad
To address the poor performance, we create a heterogeneous dataset by appending PyTorch profiles with the TensorFlow profiles. Since these two frameworks are the most prevalent, this dataset generation step is typically required only once, as it is unlikely that a new framework will significantly challenge their importance. With heterogeneous training data, Random Forest achieves 100\% Top1 accuracy, while Naive Bayes and K Nearest Neighbors exceed 95\% (see \autoref{table:heterogeneous}). This confirms that adding profiles from multiple frameworks can effectively remove the dependency on a specific framework and, therefore, ensure generalization across frameworks.
\section{Proposed Countermeasures}
\label{sec:counter}
To defend against side-channel architecture extraction attacks like InferNet, our recommendations to disrupt or mitigate the attack surface include:

\noindent
\textit{Dummy Computations:}\quad
Adding dummy GPU workload---such as no-op kernels or randomized matrix operations---can disrupt the signal-to-noise ratio in the side-channel data~\cite{cache_security_analysis, hear_shape, hermes, dnn_fingerprinting, deepsniffer, deep_peep}. Such computations should vary across inference runs to prevent adversaries from filtering them out via differential analysis. This defense impacts various GPU kernel features, such as average runtime, total runtime, and kernel invocation counts. Therefore, it requires careful design, as perturbing some features (\eg average time spent per kernel invocation) can incur significant performance overhead.

\noindent
\textit{Randomized Kernel Scheduling:}\quad
Introducing randomized scheduling between dependent or independent GPU kernels can distort execution time distributions without altering model behavior. Such noise makes it harder for adversaries to associate specific runtime patterns with architectural components. Timing jitter and minor delay injection, implemented at the framework or driver level, can increase feature variance and reduce prediction accuracy for attacks relying on stable kernel profiles.

\noindent
\textit{Limited Precision in GPU Profiling Tools:}\quad
Restricting the precision of profiling information (\eg truncating runtime metrics or kernel counts) can reduce the value of GPU profiles to adversaries~\cite{rendered_insecure, leaky_dnn}. This approach is especially effective against statistical learning methods, which rely on feature importance derived from fine-grained runtime behavior. However, tuning the defense requires careful assessment of the impact on the legitimate use of profiling tools for debugging and performance optimization.

\noindent
\textit{Monitor and Restrict Access to Profiling Tools:}\quad
Access to tools like \texttt{nvprof} should be restricted to trusted users with administrative privileges~\cite{rendered_insecure, leaky_dnn}. Enabling administrator-only flags in profiling tools can prevent adversaries from leveraging these tools for side-channel attacks. Similarly, inference services should audit profiler invocations and implement rate-limiting or sandboxing to detect repeated or automated profiling attempts.

\section{Conclusion}
\label{sec:conclusion}
This paper presents InferNet, a novel attack method that uses simple, non-intrusive GPU performance metrics to predict the underlying DNN architecture of a victim application. Unlike traditional model extraction techniques, which rely on fine-grained data or full access to the model’s parameters, InferNet only requires high-level system metrics such as GPU kernel calls, memory usage, and clock speed, making it both practical and stealthy. Our empirical evaluation demonstrates that InferNet achieves high accuracy across a diverse set of DNN architectures, including vision models, large language models, and transformers, with minimal data and even under model compression or optimization. The method is robust across multiple AI/ML frameworks and GPU platforms, providing a powerful tool for identifying DNN architectures without requiring root access or detailed model parameters.

\bibliographystyle{ACM-Reference-Format}
\bibliography{bibliography}

\appendix

\section{Model List}
\label{app:models}
\begin{table}[h!]
\caption{The 56 TorchVision and 4 TorchText architectures that are part of our InferNet attack.}
\begin{tabular}{@{}lll@{}}
\toprule
AlexNet     & Inception\_v3    & GoogleNet   \\
ConvNeXt\_Tiny     & ShuffleNet\_v2\_x0\_5    & EfficientNet\_B0   \\
ConvNeXt\_Small    & ShuffleNet\_v2\_x1\_0    & EfficientNet\_B1   \\
ConvNeXt\_Base     & ShuffleNet\_v2\_x1\_5    & EfficientNet\_B2   \\
SqueezeNet1\_0     & ShuffleNet\_v2\_x2\_0    & EfficientNet\_B3   \\
SqueezeNet1\_1     & EfficientNetV2-S        & EfficientNet\_B4   \\
ViT\_b\_16         & EfficientNetV2-M        & EfficientNet\_B5   \\
ViT\_b\_32         & EfficientNetV2-L        & EfficientNet\_B6   \\
ViT\_l\_16         & VGG11                   & EfficientNet\_B7   \\
ViT\_l\_32         & VGG11\_bn              & MobileNet\_v2             \\
ResNet18           & VGG13                  & MobileNet\_v3\_large         \\
ResNet34           & VGG13\_bn              & MobileNet\_v3\_small   \\
ResNet50           & VGG16                  & MnasNet0\_5           \\
ResNet101          & VGG16\_bn          & MnasNet0\_75          \\
ResNet152          & VGG19                  & MnasNet1\_0  \\ 
ResNext50\_32x4d   & VGG19\_bn              & MnasNet1\_3           \\
ResNext101\_32x8d   & DenseNet121   &   MaxVit\_t         \\ 
Wide\_ResNet50\_2  & DenseNet169    & DenseNet161 \\
Wide\_ResNet101\_2  & DenseNet201    &  XLMR\_BASE\\
RoBERTa\_BASE  & RoBERTa\_LARGE    &  XLMR\_LARGE\\\bottomrule
\end{tabular}
\end{table}
\begin{table}
\caption{The 24 vision and 6 LLM architectures supported by TensorFlow that are part of our InferNet attack.}
\begin{tabular}{@{}lll@{}}
\toprule
ConvNeXtTiny     & DenseNet121     & EfficientNet\_B0   \\
ConvNeXtSmall    & DenseNet169    & EfficientNet\_B1   \\
ConvNeXtBase     & DenseNet201    & EfficientNet\_B2   \\
ResNet50     & MobileNetV2    & EfficientNet\_B3   \\
ResNet101     & MobileNetV3Large        & EfficientNet\_B4   \\
ResNet152         & MobileNetV3Small        & EfficientNet\_B5   \\
VGG16         & InceptionResNetV2        & EfficientNet\_B6   \\
VGG19         & InceptionV3                   & EfficientNet\_B7   \\
XLM-RoBERTa   & GPT2                   & Gemma   \\
BERT         & BART                   & ALBERT   \\
\bottomrule
\end{tabular}
\end{table}

\section{Implementation Details and Hyperparameters}
\label{app:hyper}
We used a common preprocessing pipeline across all prediction models: feature-wise mean imputation, binary indicator augmentation for sparse kernels/events, z-score normalization, and stratified train/test splitting by architecture label. For sparse-feature experiments, features were ranked with RFE and truncated to the top-3, top-25, or full 450 non-memory GPU-kernel features, depending on the experiment. Tables~\ref{tab:infernet_setup_models}--\ref{tab:infernet_setup_metrics} summarize the experimental settings used for the architecture-prediction component of InferNet.

\noindent\textbf{Note.}
We report only the settings relevant to the architecture-prediction results in the main paper. Broader end-to-end extraction settings not directly used in the reported architecture-inference evaluation are omitted for clarity.
\begin{table*}[t]
\centering
\caption{Candidate models and execution platforms.}
\label{tab:infernet_setup_models}
\small
\begin{tabular}{p{3.0cm} p{7.5cm}}
\toprule
\textbf{Parameter} & \textbf{Value} \\
\midrule
Candidate architectures & 90 total architectures across PyTorch and TensorFlow \\
PyTorch models & 56 vision models + 4 NLP models \\
TensorFlow models & 24 vision models + 6 LLMs \\
AI/ML frameworks & PyTorch 2.3.0, TorchVision 0.18.0, TorchText 0.18.0, TensorFlow 2.17.1 \\
Profiling GPUs & NVIDIA Tesla T4, NVIDIA Quadro RTX 8000 \\
Runtime environment & Linux, Python 3.9, CUDA with \texttt{nvprof} in aggregate mode \\
\bottomrule
\end{tabular}
\end{table*}

\begin{table}[t]
\centering
\caption{GPU profile collection settings.}
\label{tab:infernet_setup_profiles}
\small
\begin{tabular}{p{3.0cm} p{7.5cm}}
\toprule
\textbf{Parameter} & \textbf{Value} \\
\midrule
Inferences per profile & 10 \\
Profiles per architecture & 20 \\
Input configuration & Single-input inference; no batch execution \\
Per-kernel statistics & \texttt{min\_us}, \texttt{max\_ms}, \texttt{avg\_us}, \texttt{num\_calls}, \texttt{time\_ms}, \texttt{time\_percent} \\
System signals & SM clock, memory clock, temperature, power, fan speed \\
\bottomrule
\end{tabular}
\end{table}

\begin{table}[t]
\centering
\caption{Preprocessing and feature-engineering settings.}
\label{tab:infernet_setup_features}
\small
\begin{tabular}{p{3.0cm} p{7.5cm}}
\toprule
\textbf{Parameter} & \textbf{Value} \\
\midrule
Train/test split & Stratified 75\% / 25\% by architecture label \\
Indicator features & Binary presence/absence indicators for GPU kernels and events \\
Missing-value handling & Column-wise mean imputation \\
Feature normalization & Standardization followed by min-max normalization \\
Feature selection & Recursive Feature Elimination (RFE) \\
Baseline feature subsets & All (849), System (27), No-system (822), GPU-kernel (447), API-call (375), Indicator (68), No-indicator (781) \\
Reduced kernel subsets & Top-3, Top-25, and all 450 non-memory GPU-kernel features \\
Transfer-study subset & Top-25 non-memory GPU-kernel features \\
\bottomrule
\end{tabular}
\end{table}

\begin{table}[t]
\centering
\caption{Victim-model training settings used for modified and pruned evaluations.}
\label{tab:infernet_setup_training}
\small
\begin{tabular}{p{3.0cm} p{7.5cm}}
\toprule
\textbf{Parameter} & \textbf{Value} \\
\midrule
Dataset & CIFAR-10 for modified/pruned victim-model generation \\
Base models & ImageNet-pretrained models where applicable \\
Optimizer & SGD \\
Default learning rate & 0.1 \\
Momentum & 0.9 \\
Nesterov momentum & Enabled \\
Weight decay & $10^{-4}$ \\
Loss function & Cross-entropy loss \\
LR scheduler & ReduceLROnPlateau \\
Scheduler patience & 10 \\
Epochs & 150 \\
Batch size & 128 \\
Workers & 8 \\
Overrides & AlexNet, ResNeXt50\_32x4d, ResNeXt101\_32x8d, and all VGG variants use LR = 0.01; all MNASNet variants use LR = 0.001; SqueezeNet1\_0 and SqueezeNet1\_1 use Adam with LR = $10^{-4}$ \\
\bottomrule
\end{tabular}
\end{table}

\begin{table}[t]
\centering
\caption{Pruning and modified-model evaluation settings.}
\label{tab:infernet_setup_pruning}
\small
\begin{tabular}{p{3.0cm} p{7.5cm}}
\toprule
\textbf{Parameter} & \textbf{Value} \\
\midrule
Pruning method & Unstructured L1 pruning \\
Pruning ratio & 0.5 \\
Fine-tuning epochs & 20 \\
Modified-model setting & Base ImageNet models (1000 classes) adapted to CIFAR-10 (10 classes) \\
\bottomrule
\end{tabular}
\end{table}

\begin{table}[t]
\centering
\caption{Reported evaluation metrics.}
\label{tab:infernet_setup_metrics}
\small
\begin{tabular}{p{3.0cm} p{7.5cm}}
\toprule
\textbf{Parameter} & \textbf{Value} \\
\midrule
Primary metrics & Train accuracy, test accuracy, Top-1 accuracy \\
Additional metrics & Top-5 accuracy, architecture-family accuracy \\
Ranking outputs & Top-$k$ accuracy and Top-$k$ confidence rankings \\
\bottomrule
\end{tabular}
\end{table}

\end{document}